\documentclass[runningheads,a4paper, article]{llncs}
\usepackage{makeidx}% allows for indexgeneration
\usepackage{graphicx}
\usepackage{subcaption}
\usepackage[justification=centering]{caption}
\usepackage{amsmath}
\usepackage{algorithm}
\usepackage[noend]{algpseudocode}
\makeatletter
\def\BState{\State\hskip-\ALG@thistlm}
\makeatother

\usepackage{algpseudocode,algorithm,algorithmicx}

\newcommand*\Let[2]{\State #1 $\gets$ #2}
\algrenewcommand\algorithmicrequire{\textbf{Precondition:}}
\algrenewcommand\algorithmicensure{\textbf{Postcondition:}}

\begin{document}

\title{Improving grasp performance using in-hand proximity and contact sensing}

% \author{Ivar Ekeland\inst{1} \and Roger Temam\inst{2}
% Jeffrey Dean \and David Grove \and Craig Chambers \and Kim~B.~Bruce \and
% Elsa Bertino}

\author{Radhen Patel, Rebecca Cox, Branden Romero and Nikolaus Correll}

% \authorrunning{Ivar Ekeland et al.} % abbreviated author list (for running head)

% \tocauthor{Ivar Ekeland, Roger Temam, Jeffrey Dean, David Grove,
% Craig Chambers, Kim B. Bruce, and Elisa Bertino}
%
\institute{University of Colorado Boulder, Boulder, CO 80309, USA}
% \email{I.Ekeland@princeton.edu},\\ WWW home page:
% \texttt{http://users/\homedir iekeland/web/welcome.html}
% \and
% Universit\'{e} de Paris-Sud,
% Laboratoire d'Analyse Num\'{e}rique, B\^{a}timent 425,\\
% F-91405 Orsay Cedex, France}

\maketitle

\begin{abstract}
We describe the grasping and manipulation strategy that we employed at the autonomous track of the Robotic Grasping and Manipulation Competition at IROS 2016. A salient feature of our architecture is the tight coupling between visual (Asus Xtion) and tactile perception (Robotic Materials), to reduce the uncertainty in sensing and actuation. We demonstrate the importance of tactile sensing and reactive control during the final stages of grasping using a Kinova Robotic arm. The set of tools and algorithms for object grasping presented here have been integrated into the open-source Robot Operating System (ROS).

% \keywords{Grasping and Manipulation pipeline, Tactile sensors, Reactive control.}
\end{abstract}

\section{Introduction}
Grasping and manipulation tasks are system-level problems that require tight integration of mechanism design, perception, and planning. In a nutshell, a robot has to locate an object, plan and execute a grasp, and finally apply sufficient constraints to the object so that it remains in the robots hand. If the task goes beyond simple pick-and-place and requires further manipulation of the object, the robot also needs to consider the pose of the object. Choosing a perception system, a suitable end-effector, and a feasible plan is a co-design problem that has been dramatically facilitated with the emergence of standardized platforms such as the PR2 robot, Rethink Robotics Baxter, and open-source software such as ROS, OpenCV and MoveIt! \cite{coleman14b}. Yet, only very few system-level grasping and manipulation studies exist, notably platforms presented at the Amazon Picking Challenge \cite{correll2016analysis}, the autonomous butler Herb \cite{srinivasa2010herb}, the PR2 \cite{bohren2011towards}, and other service robots that include manipulation for delivery, assembly or gardening tasks \cite{breuer2012johnny,knepper2010hierarchical,correll2010indoor}.

These studies are important, because the components of a grasping and manipulation system are difficult to benchmark in isolation. Specifically, it is often unclear exactly what assumptions have been made and how changes in these assumptions would affect the reliability and robustness of the system. At the task level, it is difficult to choose tasks that are representative for a wide range of real world manipulation tasks. For example, it is possible to score well in a pick-and-place competition by exclusively focusing on items that can be retrieved using suction. 

The First Grasping and Manipulation competition at the International Conference on Intelligent Robots and Systems challenged the community to solve a wide variety of grasping and manipulation tasks that range from simple bin-picking tasks to performing complex sequences of pick-and-place tasks. The competition rules promote general solutions by only combining scores achieved with the same hand. In this spirit, we have developed a comprehensive autonomous grasping solution around a Kinova Jaco 7-DoF robotic arm, RGB-D sensor (Asus Xtion), and a three-fingered hand (Kinova) equipped with proximity and tactiel sensors (Robotic Materials).  The resulting system combines deliberate planning with reactive control  using an intricate grasp state machine whose transitions are driven by 3D-perception and tactile events.  

\subsection{Related work}
We provide a brief overview over related work in the sub fields that comprise grasping and manipulation.

What hand mechanism design is most suitable to address a large variety of tasks remains an open question. At one end of the spectrum there are anthropomorphic hands with multiple degrees of freedom \cite{ilimb,bebionic,michelangelo}; on the other end there are simple one degree-of-freedom prehensors \cite{baxtergripper} and underactuated devices \cite{dollar2010highly} or soft robotic hands  \cite{farrow2016morphological,deimel2015novel}, which are entirely made out of soft and compliant materials or structures rather than of rigid parts. Although intuition would suggest that a robotic end-effector's versatility is related to its level of anthropomorphism, existing devices have been unable to accurately recreate the features of the human hand, making simple, easier to control designs  competitive. 

Planning for grasping and manipulation tasks has been traditionally studied using two distinct approaches: knowledge-based approaches and analytic approaches.  The former is based on empirical studies of human grasping and manipulation \cite{cutkosky1990human}, while the latter is based on physical models, that is the interactions between the hand and grasped object are modeled in terms of motions and forces, using the laws of physics \cite{miller2004graspit}. However, each approach has its own disadvantages. As the mechanical and sensorial mechanisms of the human hand are difficult to reproduce and it is yet unclear how sensing and actuation interact, knowledge-based approaches are only of limited use \cite{balasubramanian2012physical}. Also, it is not clear how to generalize human-inspired grasps for novel objects.

Although the analytic approaches may allow a robot to reason about how to grasp a certain object by itself, the abstractions made in the analysis to make it tractable results in models that often are only applicable to simulations or carefully structured laboratory experiments \cite{weisz2012pose}. 
%
%Until the year 2000 the field of robotic grasping was dominated by analytical approaches. The development of empirical (or data-driven) approaches started with the availability of grasp synthesizing simulators like Graspit! and OpenRAVE. In contrast, empirical approaches avoid the computational complexity by dealing with sampling candidate grasps from an infinite space of possibilities. However, in the beginning empirical approaches also relied on classical grasp metrices derived from analytical approaches to rank the grasps. They were limited to simulation due to easy of developing and evaluating approaches since the environment attributes could be controlled. Nonetheless empirical approaches could not resemble the more unstructured and dynamic real world well enough to transfer the methods easily \cite{balasubramanian2012physical,weisz2012pose}.
Due to the limitations of the knowledge-based and empirical approaches, machine learning as a solution to these tasks has been on the rise. Methods vary from observing how humans grasp an object and reducing the configuration space of the robot to find pre-grasp postures \cite{ciocarlie2009hand}, learning potential grasp points from 2D images \cite{saxena2008robotic}, learning via reinforcement and imitation learning \cite{kroemer2010combining}, to learning graspable and non-graspable objects via 2D and 3D features \cite{rao2010grasping}. %Data driven (or empirical) approaches are classified and reviewed in detail in \cite{bohg2014data}, based on whether they are synthesized for known, familiar and unknown objects. 
In our work, we ignore the problem of grasp generation and hard-code strategies that work well for the competition tasks and the mechanism/sensorial capabilities of our hand. 

%In the absence of a generic grasping strategy we consider robotic grasping not to be solved entirely. Robotic grippers for instance are designed for specific tasks. For example the Rethink Robotics Electric Gripper, a parallel jaw gripper for the Baxter robot, is meant for lifting payloads up to five pounds. Being a parallel jaw gripper, it is limited to grasp objects reliably which have parallel surfaces and which lie within its grasp aperture. Thus its use is widely seen in industries where similar kind of objects are manipulated occasionally. For other diverse objects, the company provides support with the Rethink Pneumatic Gripper. These are suction grippers onto which one can attach either a single vacuum cup or a multi-cup vacuum manifold. There are also grippers which are designed as assistive technologies. For instance, the Kinova Jaco arm whose fingers are designed keeping in mind the potential human-robot interaction. The 3-fingered arm provides optional use for 2 or 3-finger operation. Unlike the Baxter gripper, the Jaco fingers have high quality frictional pads that make grasping slippery objects easier. However, objects that can be easily picked up by Baxter, for example a hammer, are difficult to grasp for the Kinova Jaco arm and vice versa. The tendon like mechanism in the Kinoca Jaco arm finger design makes it too fragile of a finger for picking up a hammer.
Designing a perception system for grasping is strongly dependent on the end-effector choice, the variety of objects that need to be grasped, and on the environment the robot needs to operate in\cite{zhang2012application}. For example, whether objects will be grasped using suction or require careful alignment with a gripper impose very different requirements on perception.  Similary, methods that compute grasps based on the perceived geometry of an object might work very well for a large number of objects, but might fail with amorphous objects, for example a net of tennis balls \cite{correll2016analysis} . Finally, whether the objects are placed nicely on a table, are cluttered, or occluded, will dramatically change the difficulty of the problem. Some approaches assume complete or partial knowledge of the object to synthesize a grasp hypothesis \cite{dune2008active,marton2010general}, while others assume no prior knowledge of the object whatsoever \cite{bone2008automated}. Regardless of the underlying perception approach, grasping is unlikely to succeed when the resulting pose estimates from perception bear uncertainty. Only when execution is robust to uncertainties in sensing and actuation, can a grasp succeed with high probability. There are a number of approaches that use contact and tactile or visual feedback during grasp execution to adapt to unforeseen situations \cite{hsiao2010contact,felip2009robust}. These approaches increase robustness under uncertainty via some feedback mechanism. Such feedback can be obtained from visual, pressure, force-torque sensors, or  %Visual feedback can be the result from tracking the hand, object or both simultaneously. Similarly to the perceptual strategies sensors are also designed and employed for specific tasks. Instances where only slip detection is required pressure sensors are engaged ubiquitously. Distance measurements are universally done using 
proximity sensors \cite{hsiao2009reactive}. %There are some sensors that detect texture, but it is unclear how they can improve grasping. 
In this work, we are building up on results from \cite{patel16,romano2011human,patel2016b}, which use proximity, distance, and dynamic tactile sensing information, to detect different grasp events and increase robustness of the overall process with respect to uncertainty in 3D perception. 
%Motivated by an absence of a generic grasping strategy and failures of our simple perception system at multiple instances we aid our grasping pipeline with a reactive control from the tactile information available from an integrated force and distance sensor. We demonstrate the results of our proposed strategy by engaging the Jaco2 arm from the Kinova Robotics into every-day object manipulation and grasping activities that a human routinely performs.

\section{Task specification}
\begin{figure}
\centering
\includegraphics[width=0.495\linewidth,height=4.5cm]{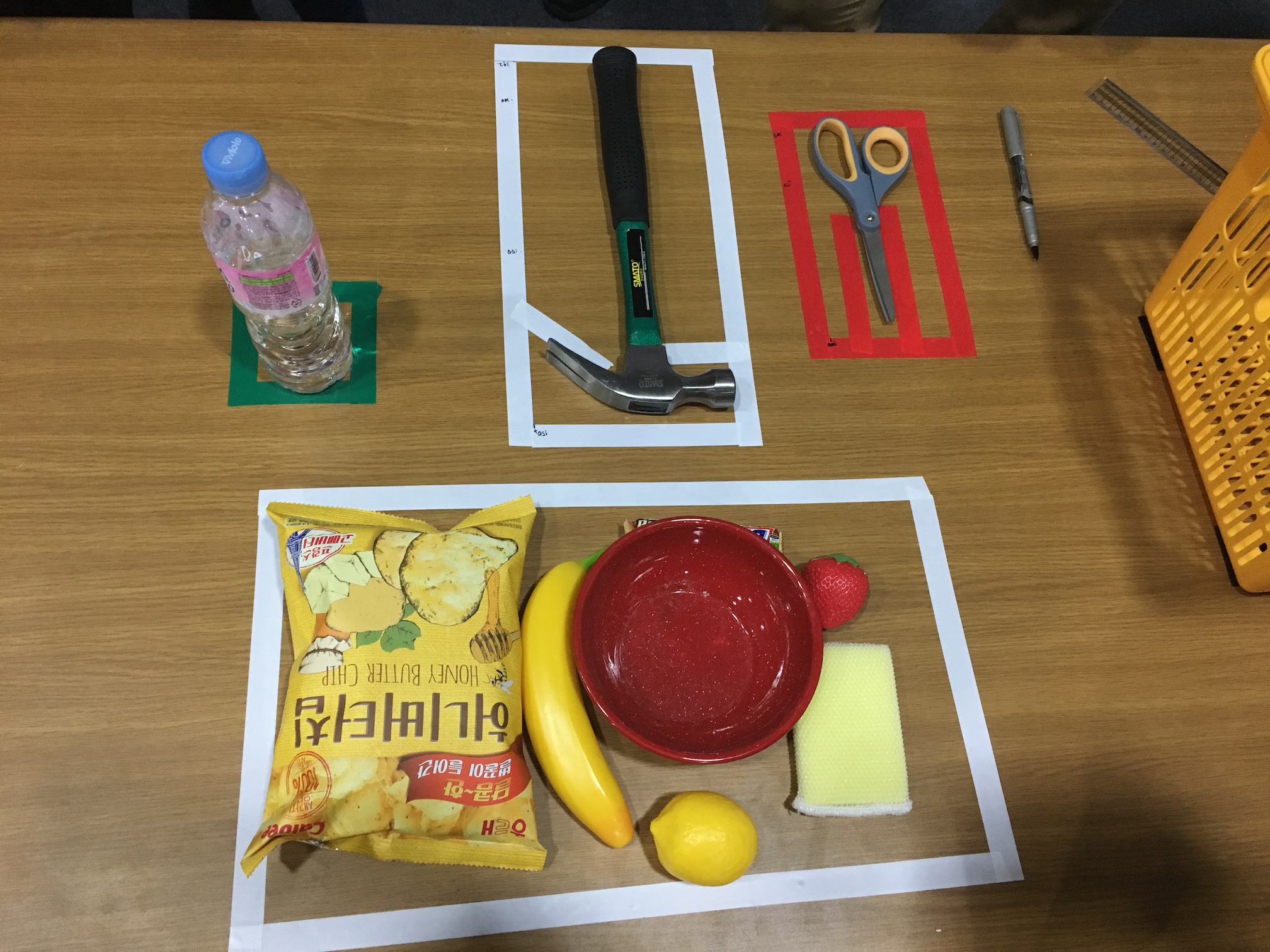}
\quad
\includegraphics[width=0.465\linewidth,height=4.5cm]{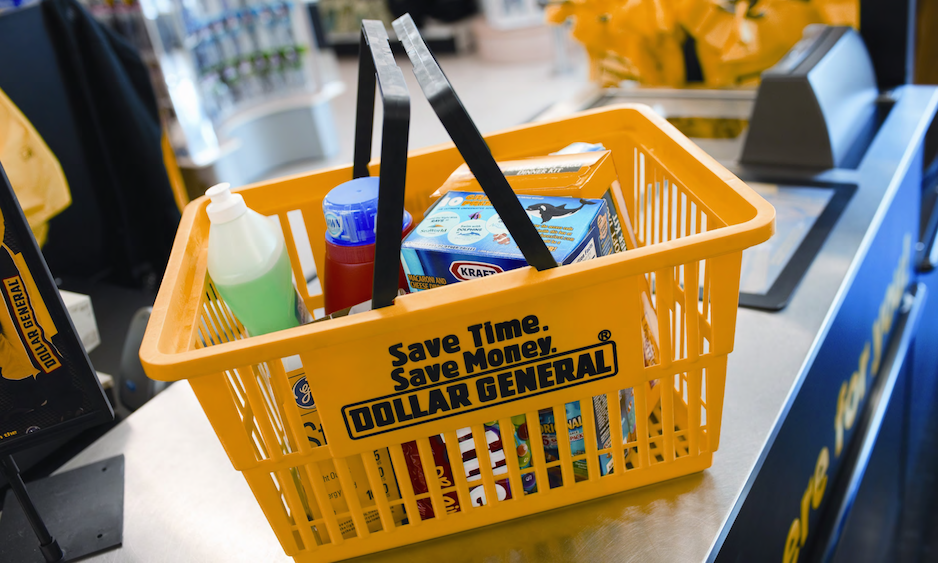}
\caption{Left: Objects and their predefined locations for the track-2 stage-1 pick and place task. Right: Basket containing all the objects.\label{fig:picknplace}}
\end{figure}

The autonomous track consisted of two stages: pick-and-place and manipulation. All sets of tasks were required to be performed fully autonomously, that is without human intervention. 
% These team were from University of Colorado at Boulder, Dorabot\&Cobot from China, Tsinghua University, Sungkyunkwan University from South Korea and George Mason University.
The pick-and-place stage required contestants to design a system that would pick and then place a set of objects into a designated area autonomously. The majority of the objects could be placed within their designated area without constraints on their orientation. A few objects had to be placed in a specific orientation, for example the hammer and the scissors as shown in Figure \ref{fig:picknplace}. The set of objects consisted of ten objects chosen from a set of twenty objects \cite{calli2015benchmarking} that were disclosed before the competition. These objects were then randomly placed within a shopping basket (Figure \ref{fig:picknplace}), and the contestant were allotted 30 minutes to perform the task. Each successful placement was rewarded five points, leading to a maximum of fifty points.

\begin{figure}[ht]
\centering
\begin{subfigure}[t]{0.24\linewidth}
\includegraphics[width=2.85cm,height=2.5cm]{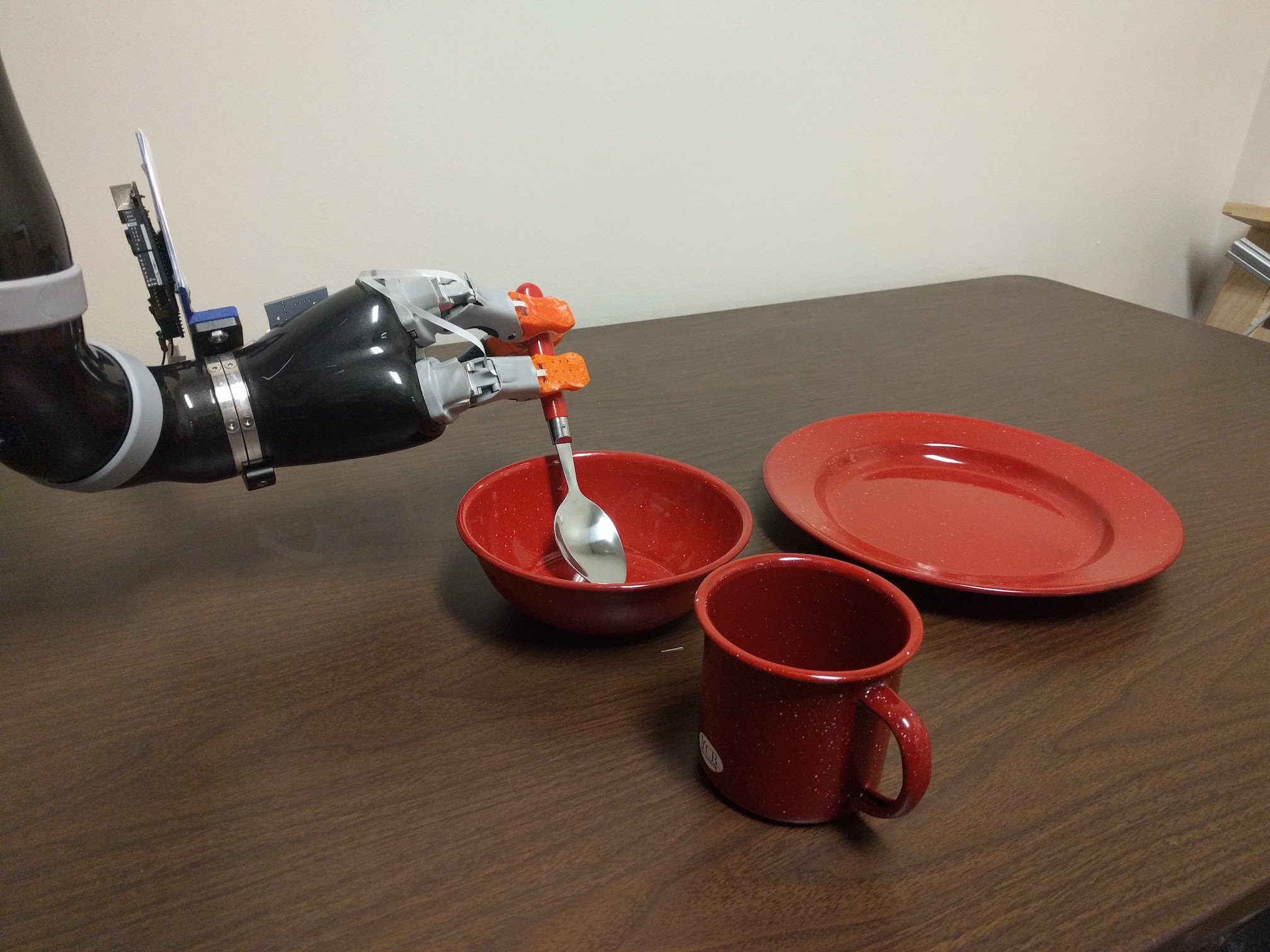}
\end{subfigure}
\begin{subfigure}[t]{0.24\linewidth}
\includegraphics[width=2.85cm,height=2.5cm]{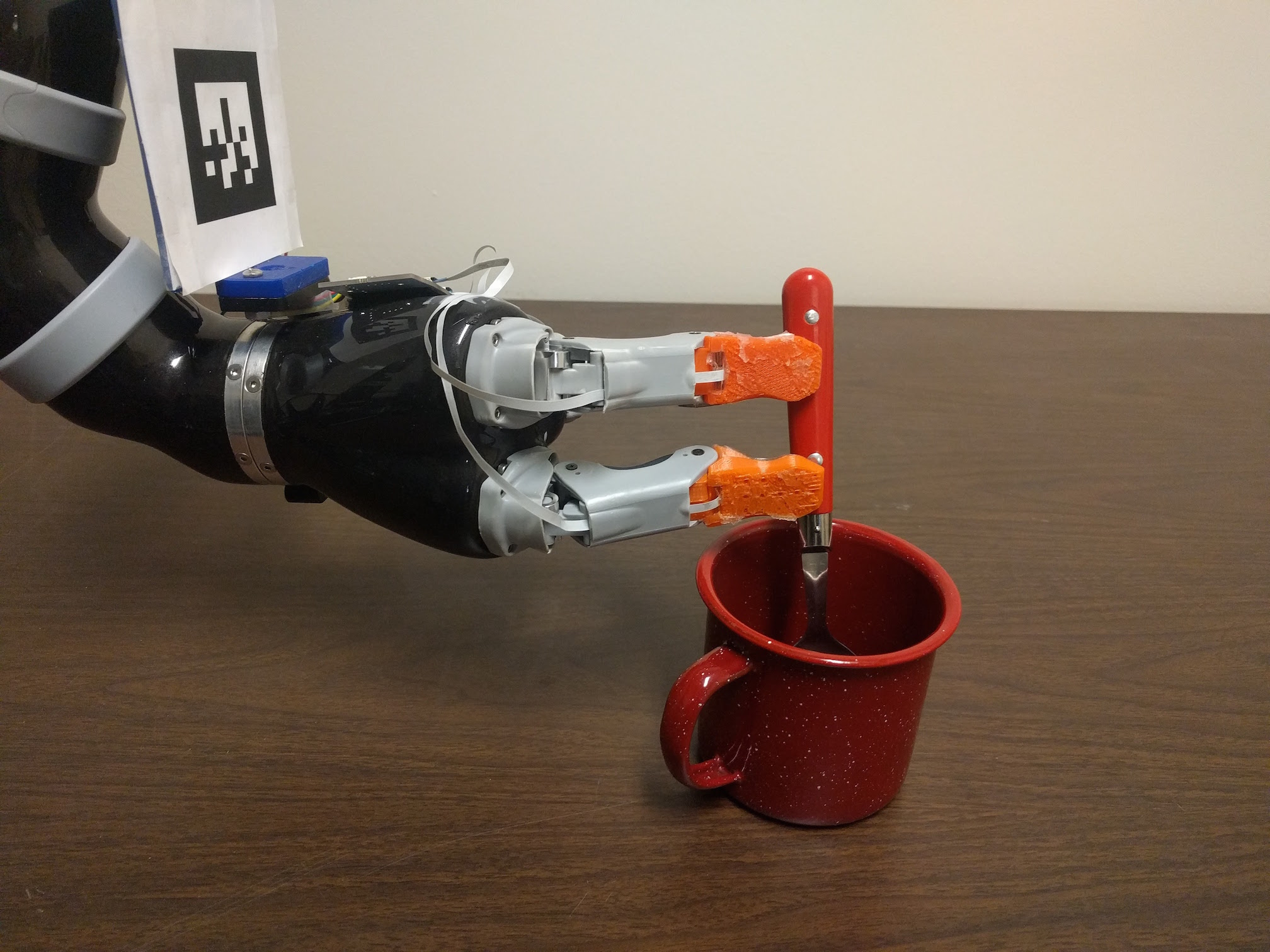}
\end{subfigure}
\begin{subfigure}[t]{0.24\linewidth}
\includegraphics[width=2.85cm,height=2.5cm]{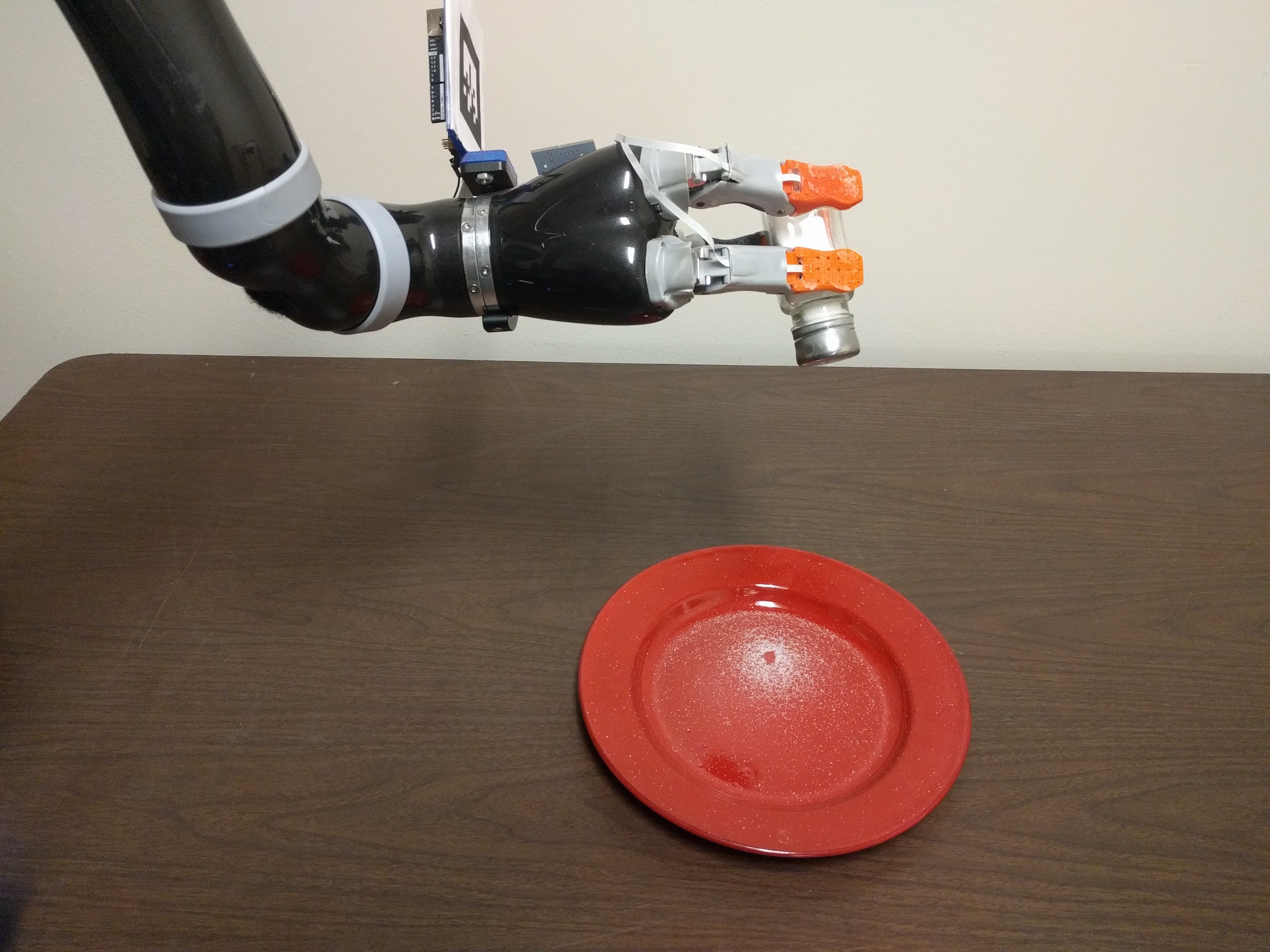}
\end{subfigure}\\\vspace{0.2em}
\begin{subfigure}[t]{0.24\linewidth}
\includegraphics[width=2.85cm,height=2.5cm]{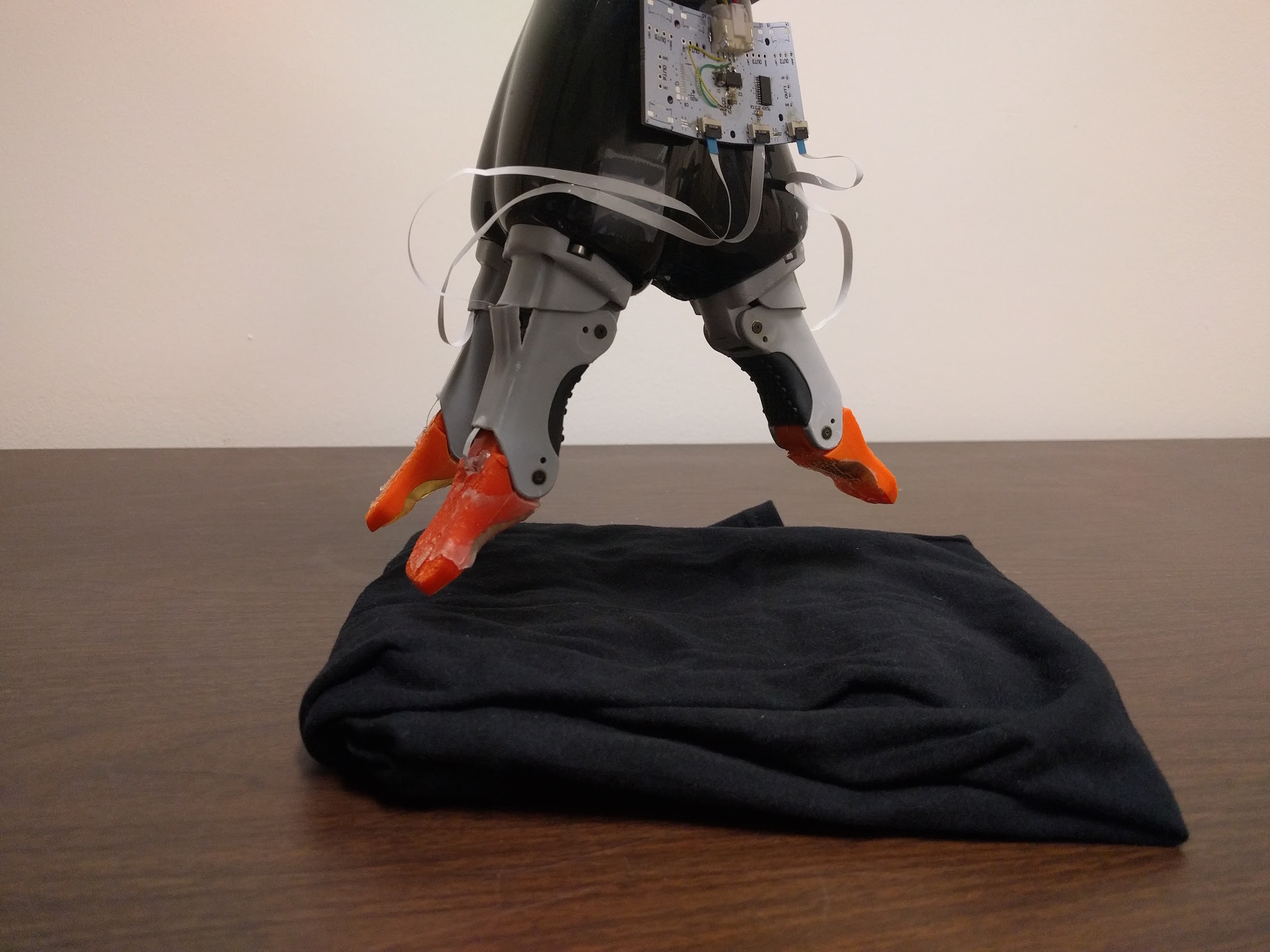}
\end{subfigure}
\begin{subfigure}[t]{0.24\linewidth}
\includegraphics[width=2.85cm,height=2.5cm]{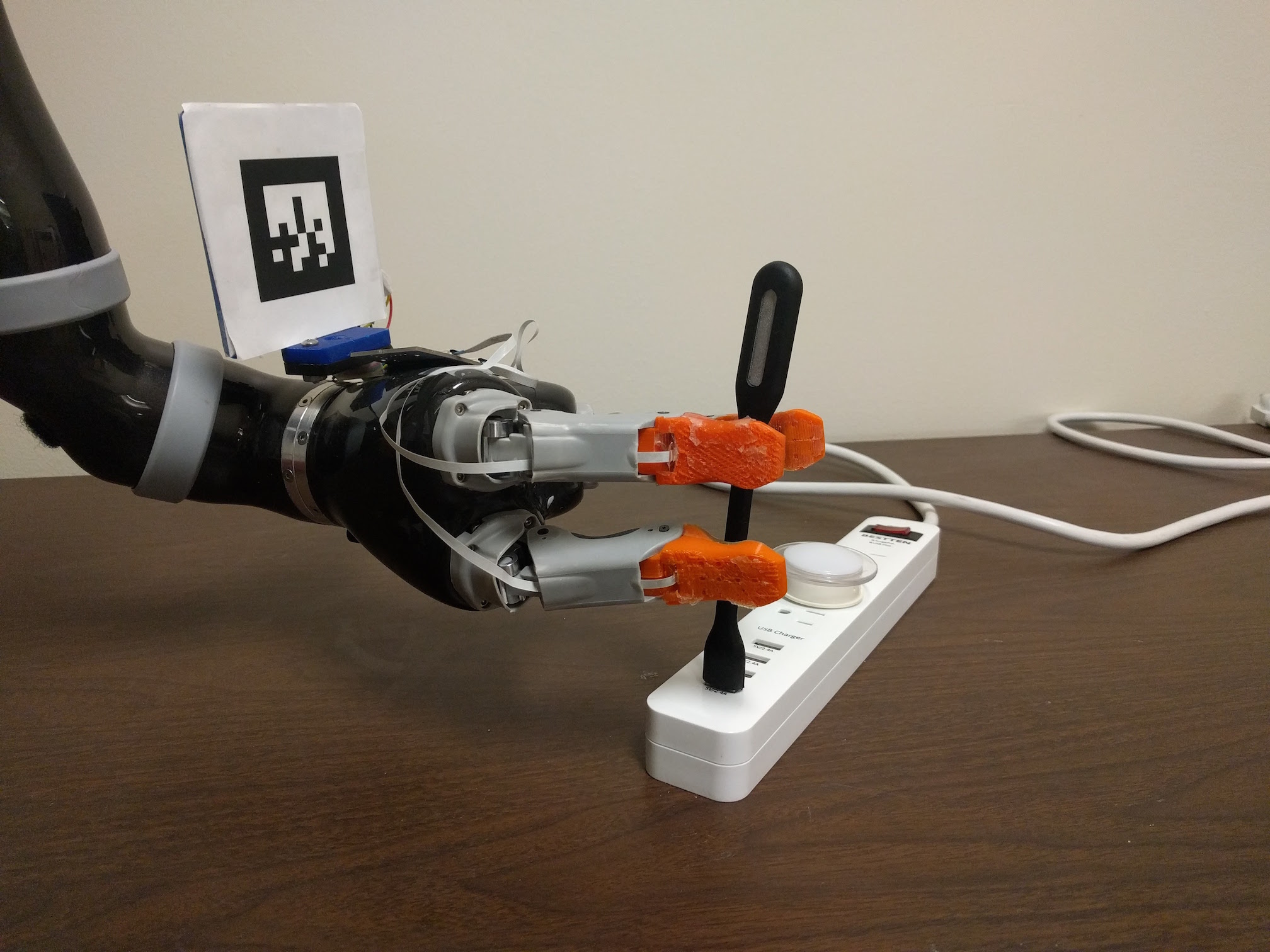}
\end{subfigure}
\begin{subfigure}[b]{0.24\linewidth}
\includegraphics[width=2.85cm,height=2.5cm]{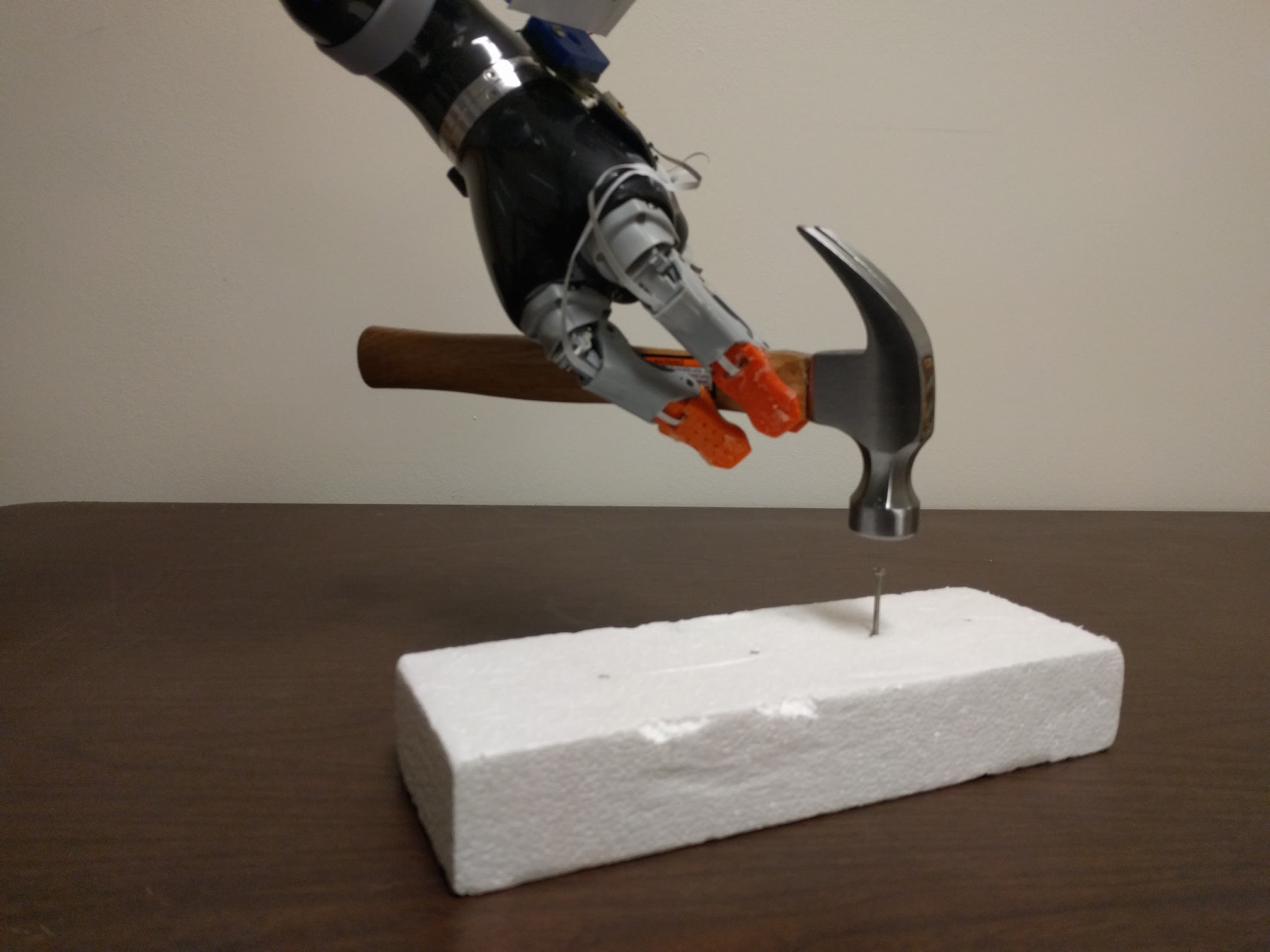}
\end{subfigure}
\begin{subfigure}[b]{0.24\linewidth}
\includegraphics[width=2.85cm,height=2.5cm]{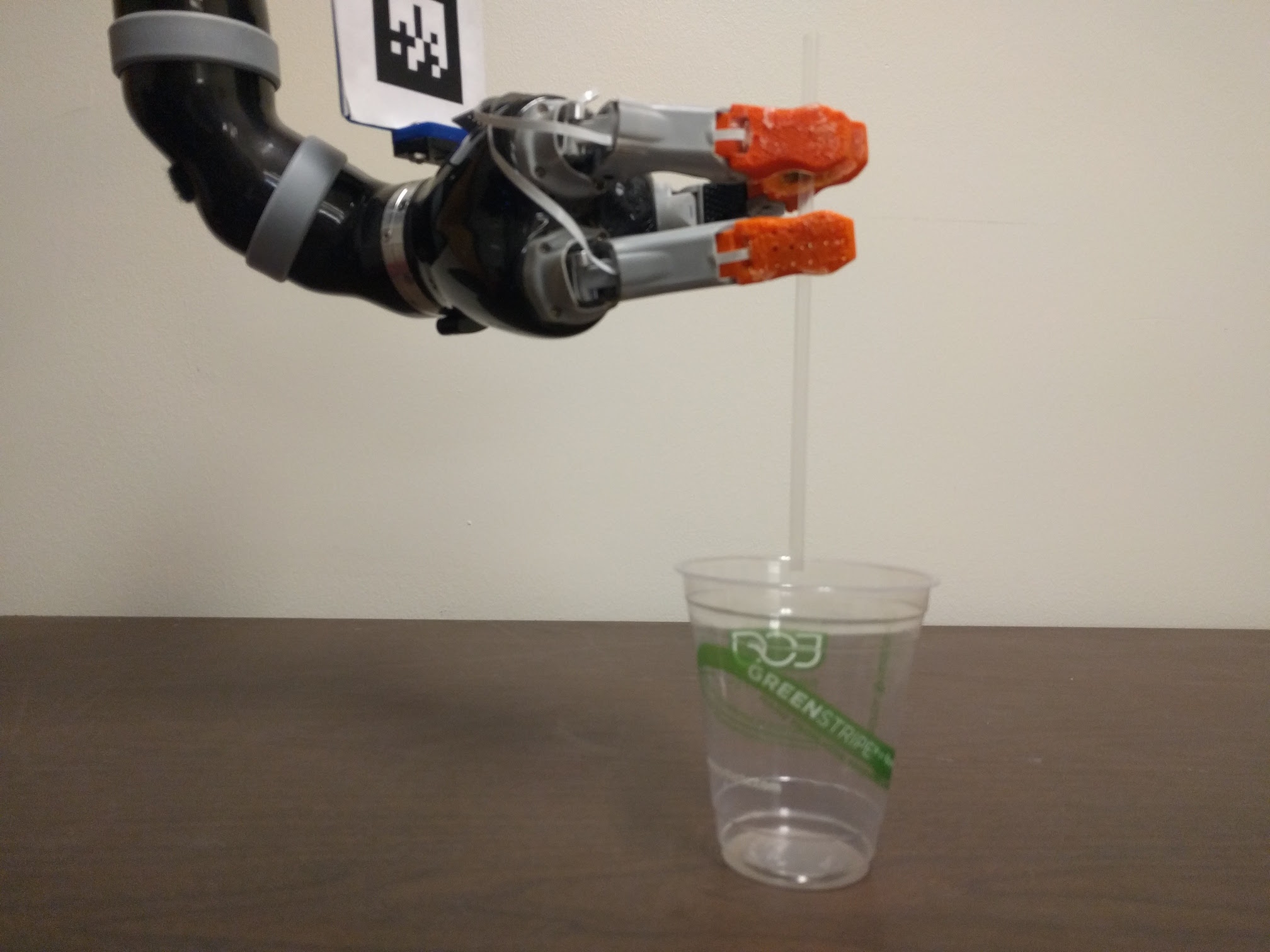}
\end{subfigure}\\\vspace{0.2em}
\begin{subfigure}[b]{0.24\linewidth}
\includegraphics[width=2.85cm,height=2.5cm]{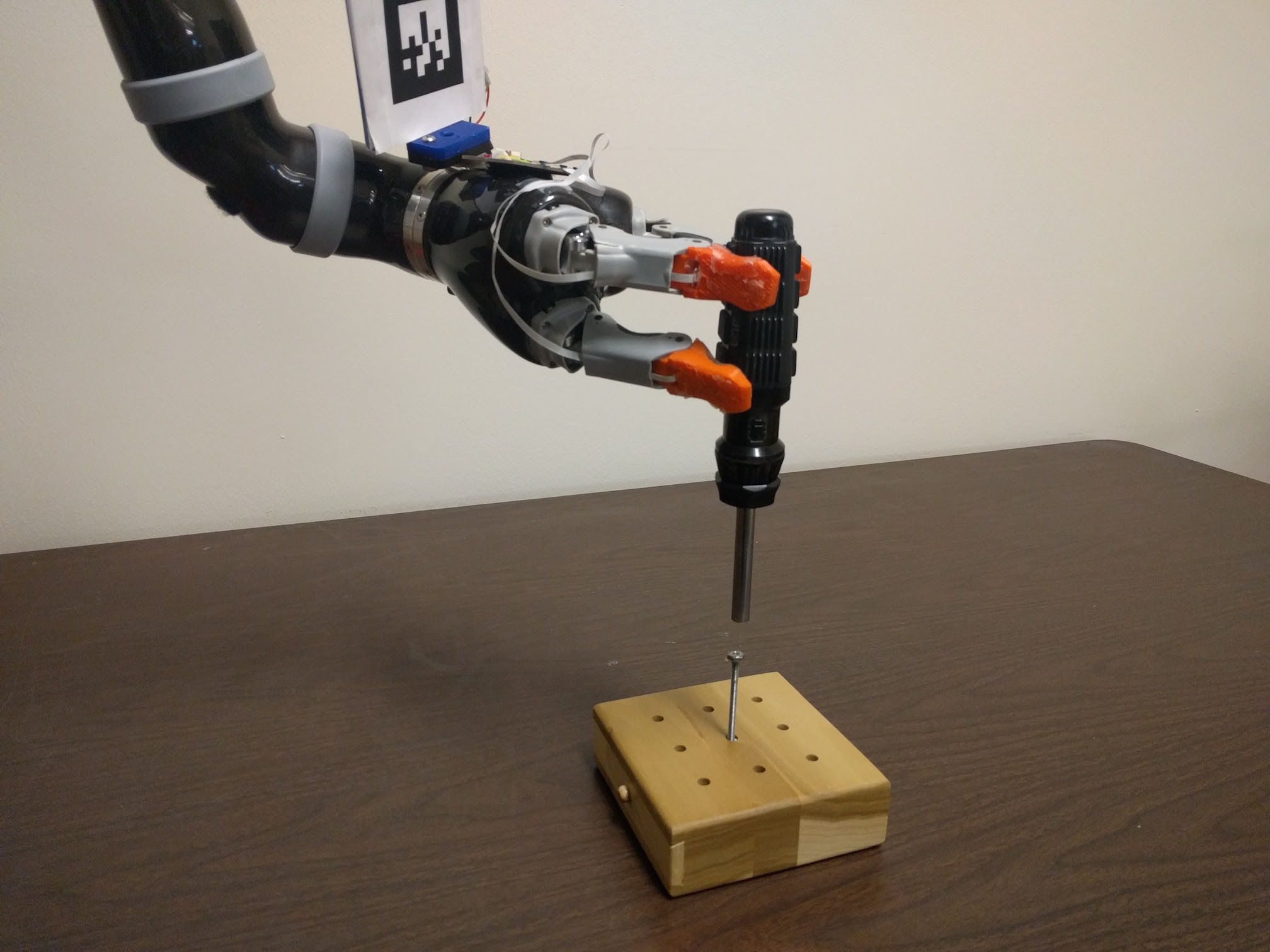}
\end{subfigure}
\begin{subfigure}[t]{0.24\linewidth}
\includegraphics[width=2.85cm,height=2.5cm]{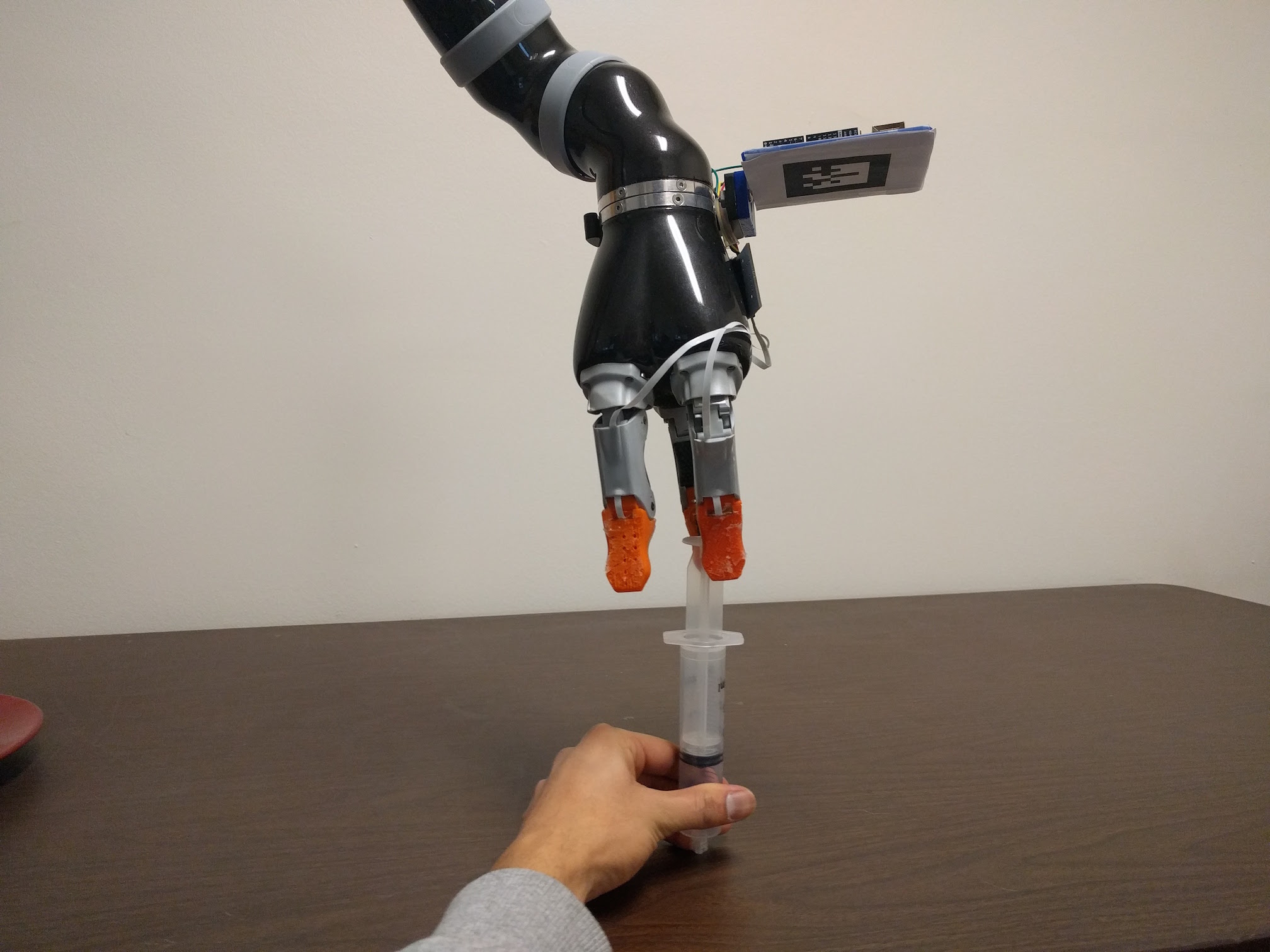}
\end{subfigure}
\begin{subfigure}[b]{0.24\linewidth}
\includegraphics[width=2.85cm,height=2.5cm]{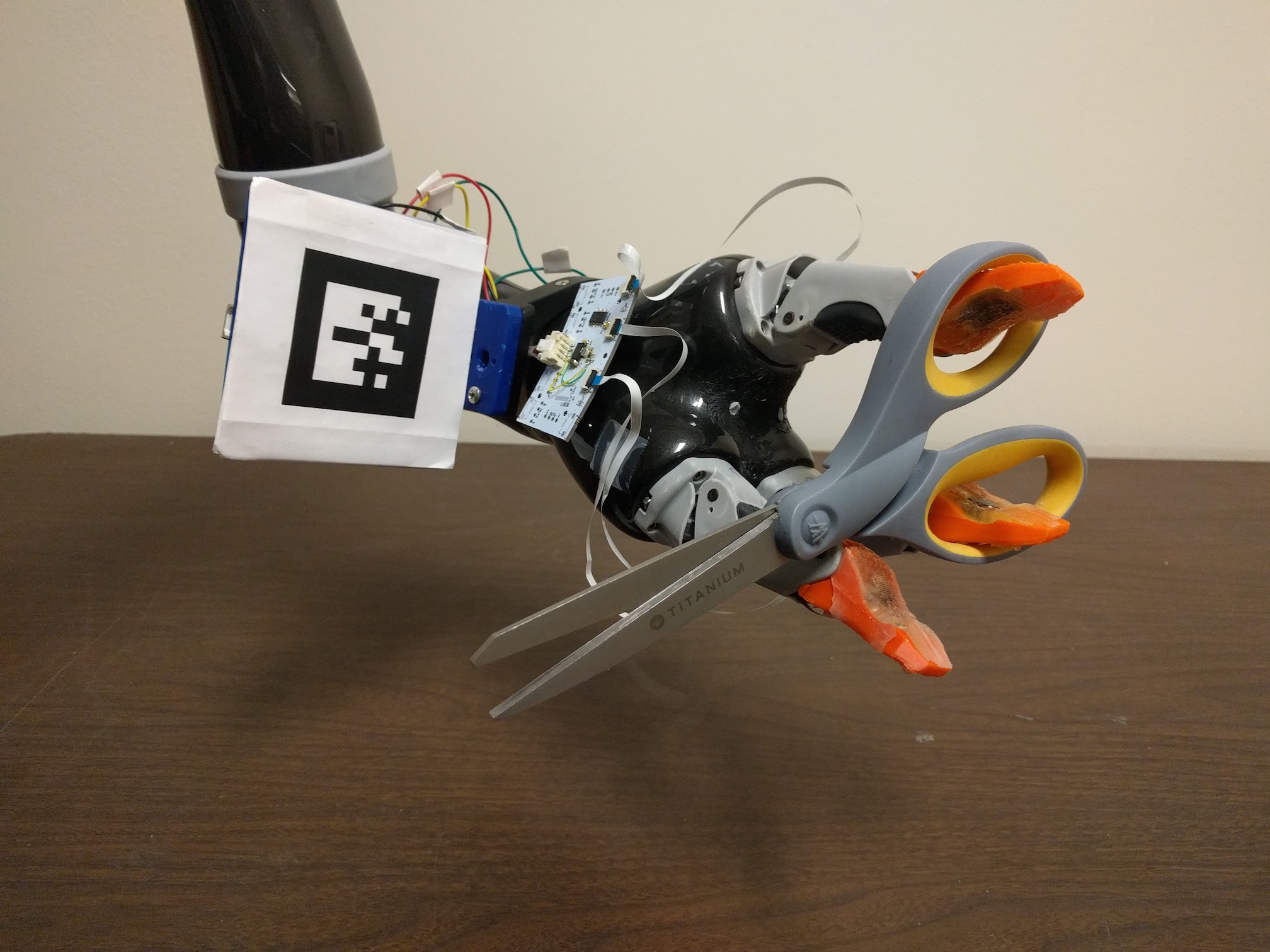}
\end{subfigure}
\caption{Manipulation tasks from the competition. Clockwise, starting top left. LEVEL1 \textit{tasks} (i) Scooping peas, (ii) Stirring, (iii) Salt shaking. LEVEL2 \textit{tasks} (iv) Towel picking, (v) Plugging and unplugging USB lights, (vi) Hammering nails, (vii) Straw inserting. LEVEL3 \textit{tasks} (viii) Nut fastening, (ix) Syringe pumping. LEVEL4 \textit{tasks} (x) Paper cutting.\label{fig:manipulation}}
\end{figure}

The manipulation stage consisted of ten tasks (Figure \ref{fig:manipulation}) that varied in difficulty. The ten tasks were selected from a pool of 18 tasks and were divided into four levels based on difficulty. Contestants that designed a system that successfully completed one of four tasks in level one were rewarded ten points, twenty points for one of three tasks in level two, thirty points for one of two tasks in level three, and forty points for the one task in level four. As a result, a maximum of 200 points could be achieved. 

% * <brro5352@colorado.edu> 2017-01-18T20:18:04.645Z:
% 
% Add picture of syringe task
% 
% ^.
% * <brro5352@colorado.edu> 2017-01-18T20:17:37.799Z:
% 
% Add to the figure what level each of the task were
% 
% ^.
\section{Technical Approach}
\begin{figure}[!htbp]
\centering
\includegraphics[width=9cm, height = 7cm, keepaspectratio]{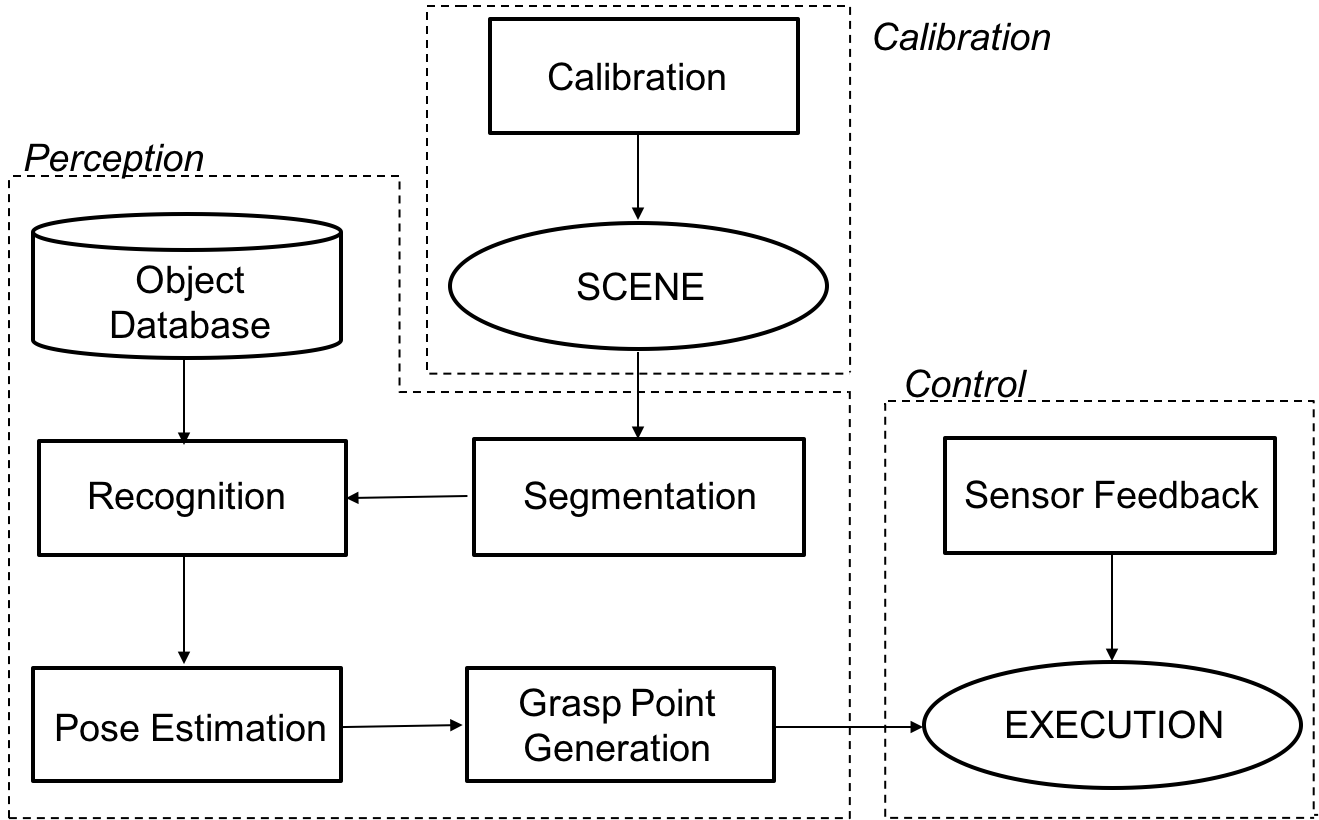}
\caption{A flow-chart depicting the various components of our system. \label{fig:statemachdiagram}}
\end{figure}

We developed a comprehensive autonomous grasping solution around a Kinova Jaco 7-DoF robotic arm, RGB-D sensor (Asus Xtion), and a Kinova three-fingered hand with proximity, contact and force sensors (Robotic Materials).  The resulting system combines deliberate planning with reactive control  using an intricate grasp state machine whose transitions are driven by 3D-perception and tactile events.  In particular, we developed a general-purpose software pipeline composed of several independent nodes that perform specific tasks such as eye-to-hand calibration, object recognition and tracking, and kinematic control and planning of the arm (Figure \ref{fig:statemachdiagram}) in the form of a Robot Operating System (ROS) package, which is available open-source\footnote{\url{https://github.com/correlllab/cu-perception-manipulation-stack}}.
% * <brro5352@colorado.edu> 2017-01-18T21:04:03.616Z:
% 
% Should we mention the visualization tools.
% 
% ^.

\subsection{Calibration} 
\begin{figure}[!htb]
\centering
\includegraphics[width=6cm, height = 3cm]{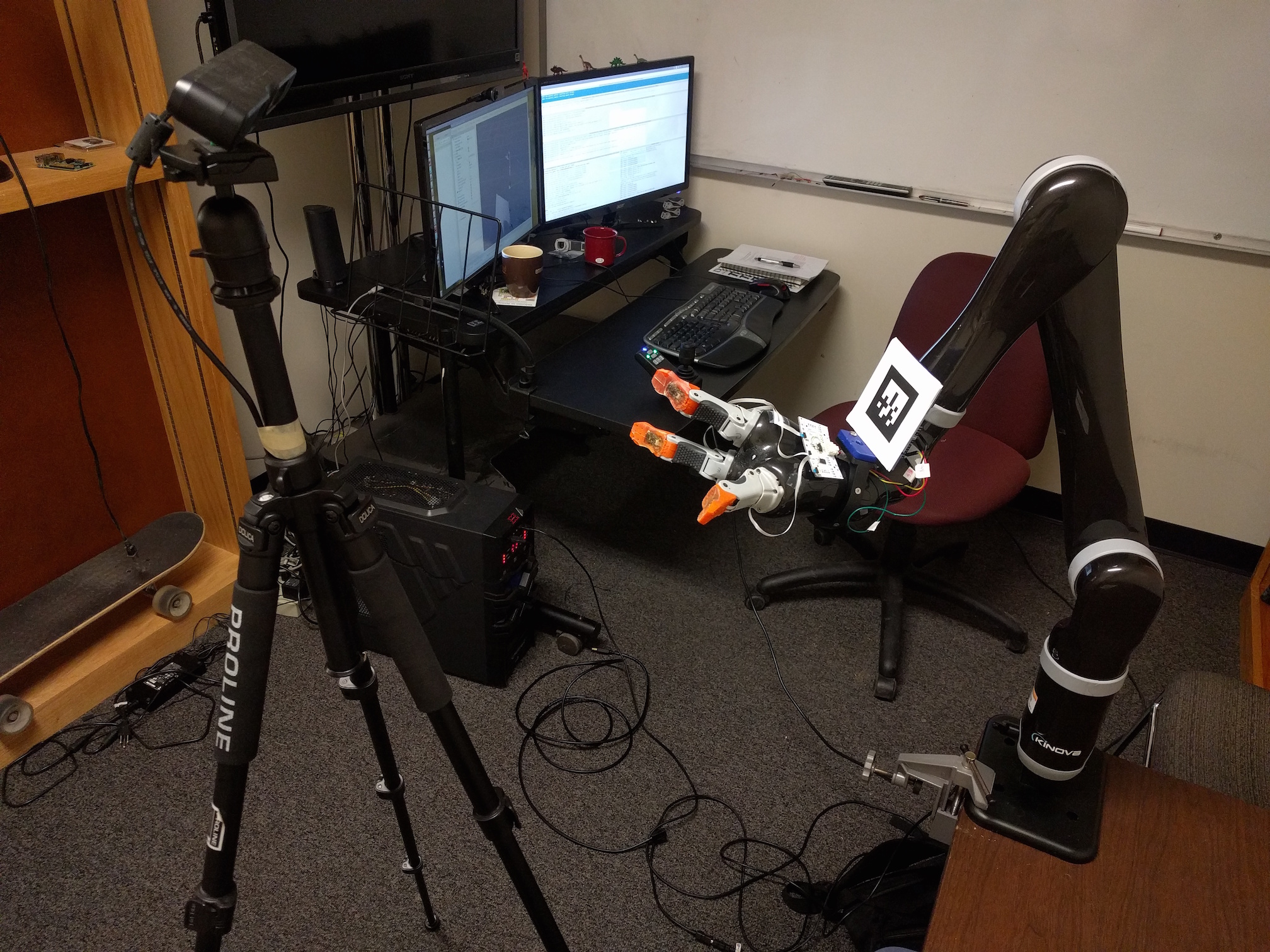}
\includegraphics[width=6cm, height = 3cm]{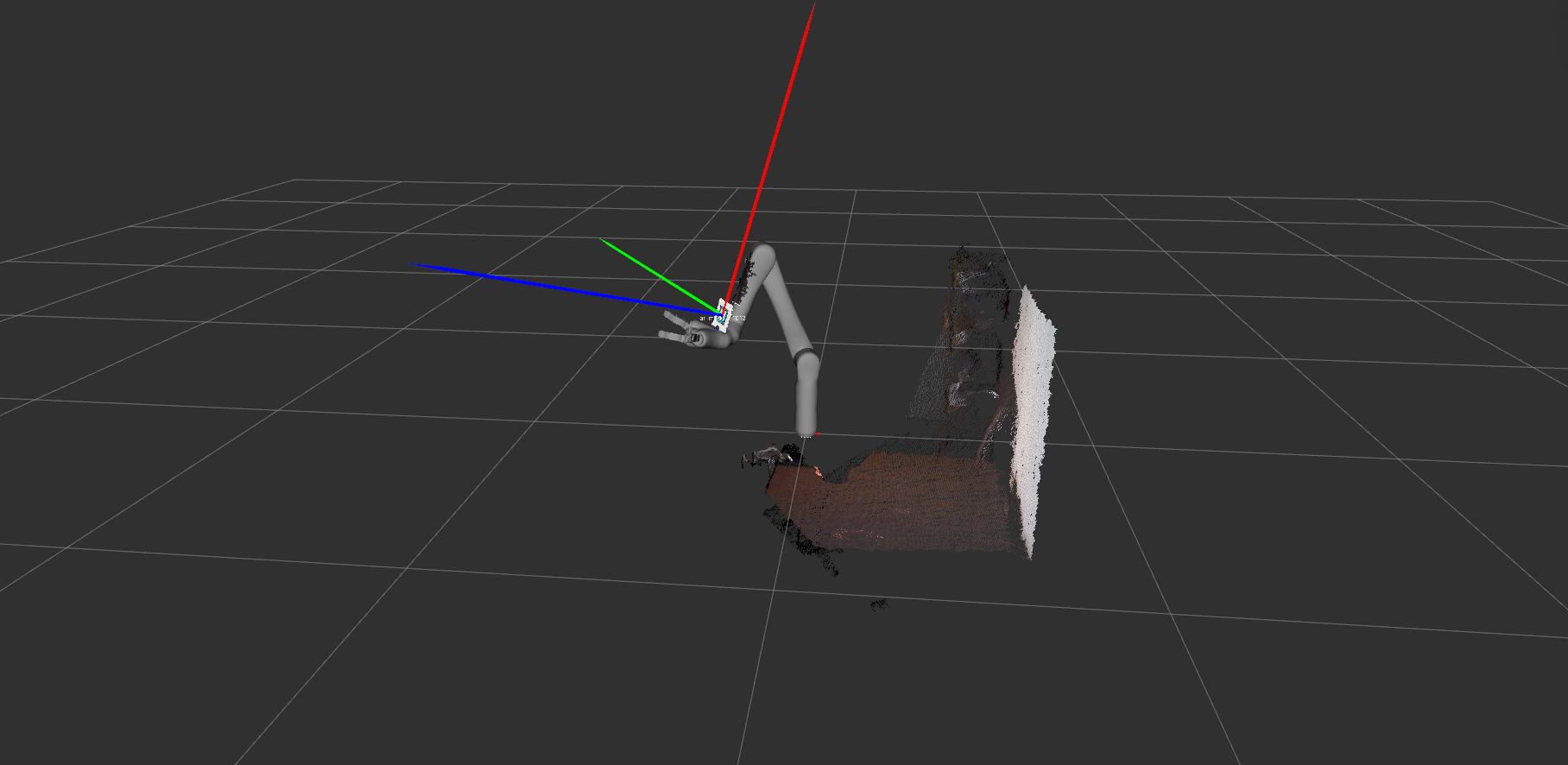}
\caption{Left: Position of camera relative to camera. Right: Model scene after calibration in RViz .\label{fig:calibration}}
\end{figure}

% * <brro5352@colorado.edu> 2017-01-18T20:57:36.199Z:
% 
% > \subsection{Calibration} 
% Should we add photos of the calibration process?
% 
% ^.
To initialize the system, the user needs to first calibrate the RGB-D camera. Our system allows the user to place the camera in a position suitable for their needs rather than rigidly attaching it to a single location. While this allows the system to quickly adapt to a variety of tasks that require different perspectives, mobility adds uncertainty to the model since the sensor's location in space is unknown. In order to find the transformation between camera and robot frame, we rigidly mounted an augmented reality (AR) tag to the wrist joint of the Jaco arm (Figure \ref{fig:calibration}, left). Once the AR tag is visible to the sensor, the system can estimate the transform between the sensor and the AR tag. Since the position of the wrist joint is known to our model, the system can then estimate the position of the sensor in space using forward kinematics. As a result, we are able to obtain a calibrated scene with an offset error of about 3 cm. The calibrated scene in RViz is shown in Figure \ref{fig:calibration}, right. 

% * <radhen.17@gmail.com> 2017-01-19T23:42:59.767Z:
% 
% > sensor
% replace sensors with camera
% 
% 
% ^.

\subsection{Perception} 
We developed a perception pipeline using the Point Cloud Library (PCL) to process the depth data received from the ASUS Xtion. In each task, the objects lie on a table or flat surface that fills a large portion of the field of view of the depth sensor. We first segment out this tabletop using a simple non-deterministic outlier detection method (RANSAC).  Filtering the tabletop out from the point cloud greatly reduces the points in our cloud and  leaves gaps between remaining objects that assist in segmentation. Using Euclidean distance, neighboring points are clustered together to form separate objects, assuming that they are sufficiently spaced apart. Objects too close together, such as a stack of blocks, are segmented using secondary features such as color. These segmented objects are then matched to already seen object templates present in the database using 3D feature detectors and labeled accordingly (e.g., cup, plate, bowl). 
% * <brro5352@colorado.edu> 2017-01-18T23:12:30.795Z:
% 
% > We first segment the table top using a simple non-deterministic outliner detection method (RANSAC)
% Maybe explain why we do this, i.e we need to segment out the table so that we have a set of point-clouds to segment if we didn't something like the objects center will be offset. 
% 
% ^.
% * <brro5352@colorado.edu> 2017-01-18T23:05:32.290Z:
% 
% > Euclidean clustering
% Maybe add a citation in case people dont know what that is. Alsso with RANSAC and PCL?
% 
% ^.
% * <brro5352@colorado.edu> 2017-01-18T23:03:49.489Z:
% 
% > simple
% Choice of vocabulary might be underselling it. Maybe something like robust perception pipeline.
% 
% ^.
% * <brro5352@colorado.edu> 2017-01-18T23:03:02.236Z:
% 
% > We first segment the table top using a simple non-deterministic outliner detection method (RANSAC). Objects on top of the percieved table top are then segmented using Euclidean clustering, assuming that they are sufficiently spaced apart
% State the assumptions made about the environment, i.e we assume that the objects are on a table top.
% 
% ^.

\begin{figure}[!htb]
\centering
\begin{subfigure}[t]{0.24\linewidth}
\includegraphics[width=2.9cm,height=3cm]{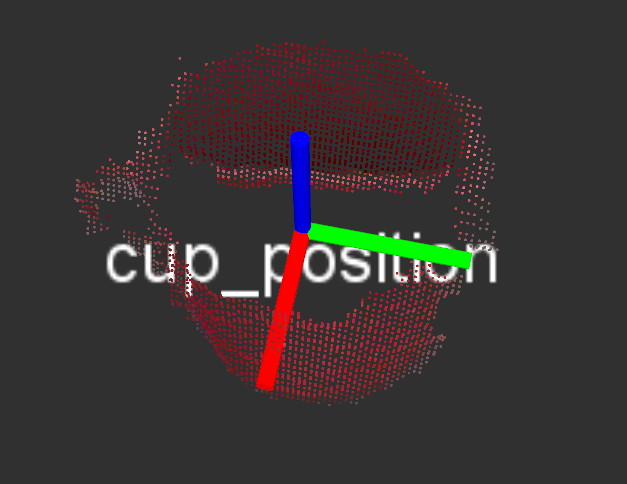}
\end{subfigure}
\begin{subfigure}[t]{0.24\linewidth}
\includegraphics[width=2.9cm,height=3cm]{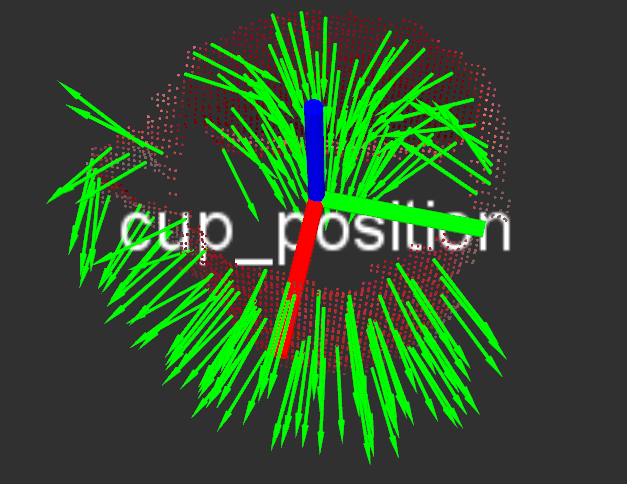}
\end{subfigure}
\begin{subfigure}[t]{0.49\linewidth}
\includegraphics[width=6cm,height=3cm]{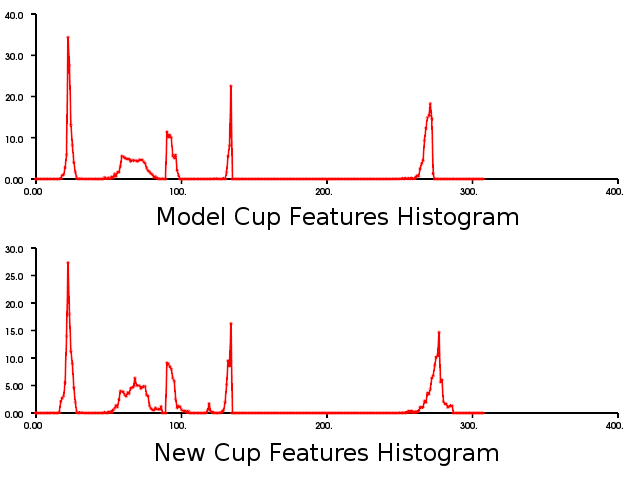}
\end{subfigure}
\caption{ 3D object Recognition of a cup. Clockwise, starting top left. (i) Point cloud of cup from the YCB Dataset, (ii) Green arrows display normals computed for a select few points in the cloud, (iii) Viewpoint Feature Histograms (VFH) showing the similarity of the model cup with the new cup \cite{rusu2010fast} \label{fig:perception}}
\end{figure}

Similar to 2D object recognition, 3D object recognition relies on finding characteristic key points and matching them to a database. Features based on the normal of a surface are reliable since it has similar values when computed for the same surface of an object in different point clouds and at different orientations. The normal of each point is calculated by taking the nearest neighbors within a defined radius to find the tangent plane. The perpendicular vector of that plane pointing towards the camera is the normal. The vector not pointing towards the camera would not be visible to the sensor, so it can clearly be discarded. An example point cloud of a cup  with computed normals and the corresponding feature histogram is shown in \ref{fig:perception}. 
% * <brro5352@colorado.edu> 2017-01-19T21:28:57.724Z:
% 
% >  XYZ coordinates are characteristic, but are not enough when interpolating the data of point clouds to find matching features.
% Why is this? Because of scaling?
% 
% ^.
% * <brro5352@colorado.edu> 2017-01-19T21:27:05.746Z:
% 
% > Similar to 2D object recognition, 3D object recognition relies on finding characteristic key points and matching them to previously seen ones. XYZ coordinates are characteristic, but are not enough when interpolating the data of point clouds to find matching features. 
% Not sure about the build up, could be more brief. 
% 
% ^.
% * <brro5352@colorado.edu> 2017-01-18T23:16:34.143Z:
% 
% > previously seen ones
% Keep the language consistent and say something like its from the database.
% 
% ^.

Next, we compare our detected features with our known database using the Signature of Histograms of Orientations (SHOT) descriptor \cite{tombari2010unique}. Histograms are computed on the orientations of normals in a sphere or 3D volume and then grouped together using their intersection to form the local descriptor. Similar to the well known SIFT algorithm for 2D object recognition, SHOT is also robust to occlusion and rotation and can be used to determine orientation. One big advantage to using 3D object recognition over 2D is the ability to use the depth data provided from the camera for estimating the location. This additional location information is used to calculate grasping orientations and for avoiding collisions. Once the camera location is found relative to the robot arm, we are able to do a simple transformation to get the object's pose relative to the robot for grasping and manipulation described later. All parameters of our processing pipeline are accessible in a user interface, allowing us to fine tune parameters to lighting conditions and changes in camera pose in the competition environment.
% * <brro5352@colorado.edu> 2017-01-19T21:32:52.712Z:
% 
% > One big advantage to using 3D object recognition over 2D is the ability to estimate the location of the object relative to the sensor.
% Maybe just mention that from this info we can estimate the pose, but with this can't we get the estimate of the location using just depth data/ SHOT is for 3d object recognition not position estimation, and then once the object is recognized we use depth data to estimate the position.
% 
% ^.
% * <brro5352@colorado.edu> 2017-01-19T21:31:31.071Z:
% 
% > re computed over all the points in a sphere or 3D volume and then grouped together using their intersection to form the descriptor
% Confused about this, I thought you were using normals for this. If not, where do the normals from the previous paragraph fit into the object detector
% 
% ^.
% * <brro5352@colorado.edu> 2017-01-19T21:30:11.141Z:
% 
% > these
% Might want to say SHOT, not clear what "these" are from context
% 
% ^.

\begin{figure}[!htb]
% * <radhen.17@gmail.com> 2017-01-21T18:16:56.391Z:
% 
% > \begin{figure}[!htb]
% > \centering
% > \includegraphics[width=6cm, height = 4cm]{figs/SYS_1.png}
% > \quad
% > \includegraphics[width=5.5cm, height = 4cm]{figs/SYS_2.jpg}
% > \caption{Left: Calibrated view of the experimental setup and the Jaco2 arm as seen in RViz. Right: Custom made fingers and integrated proximity and tactile sensors on Jaco2 arm.\label{fig:setup}}
% > \end{figure} 
% if we dont keep the uncalibrated view (the third image) in Fig4 then this and that image are almost similar.
% we should keep any one?
% 
% ^.
\centering
\includegraphics[width=6cm, height = 4cm]{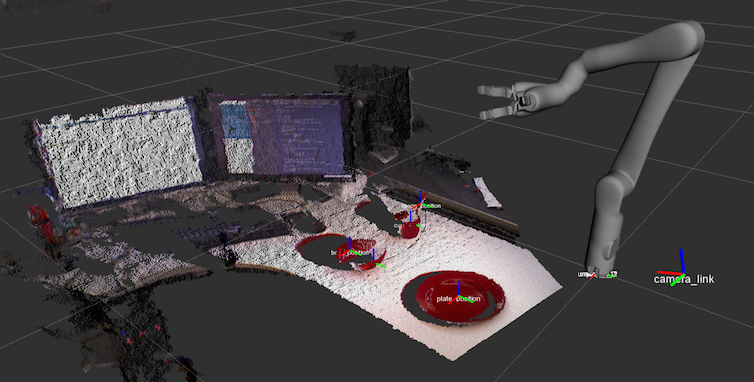}
\quad
\includegraphics[width=5.5cm, height = 4cm]{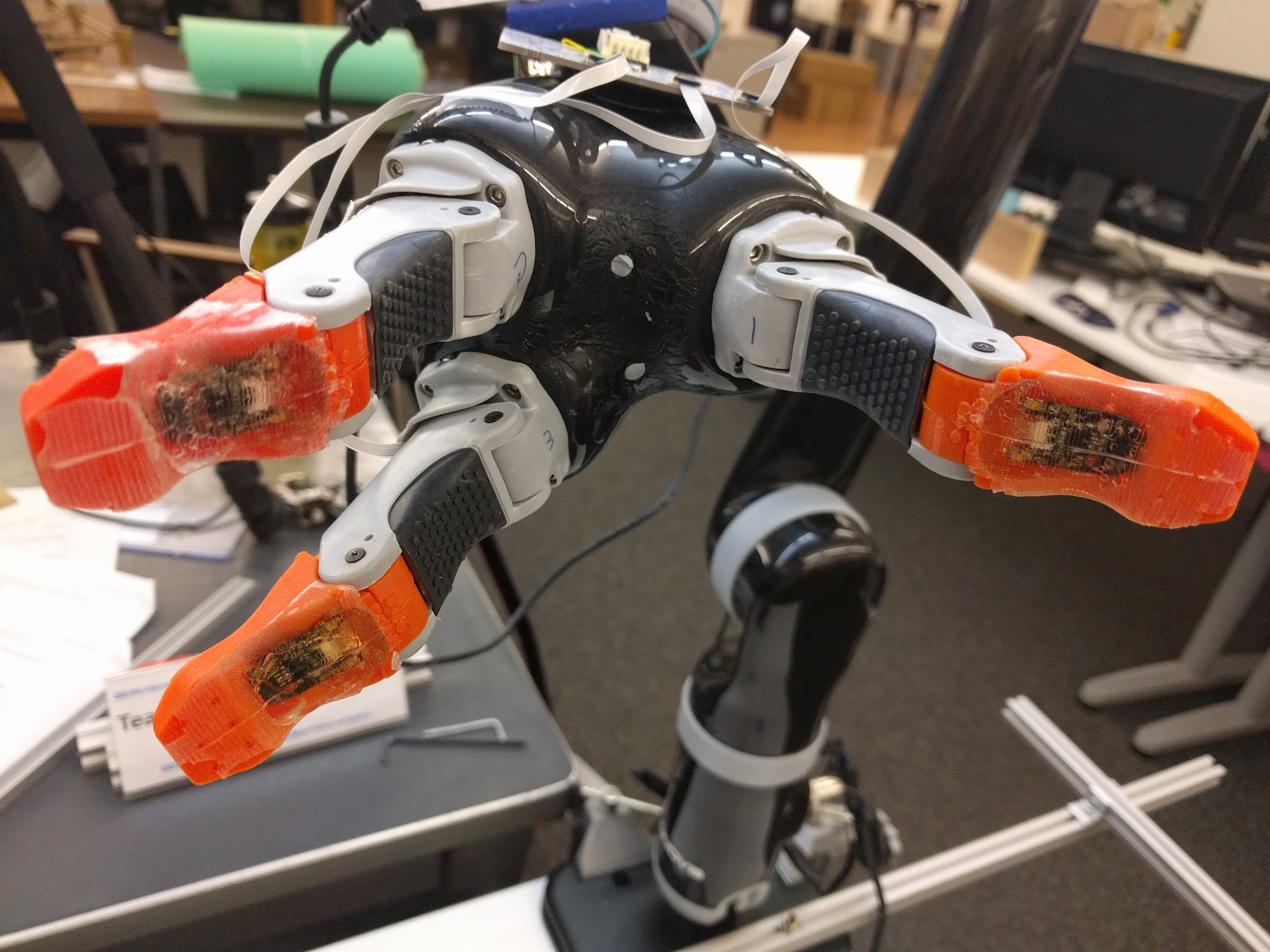}
\caption{Left: Calibrated view of the experimental setup and the Jaco2 arm as seen in RViz. Right: Custom made fingers and integrated proximity and tactile sensors on Jaco2 arm.\label{fig:setup}}
\end{figure} 

\subsection{Control} 
The control node of our system controls the arm through two modes, Cartesian control and velocity control. The mode chosen at a particular time step depends on the action being executed. In particular, we split up control into two distinctive actions, approach and search. The former deals with large scale movements that put the end-effector in the vicinity of the object of interest, while the latter uses the feedback from the finger sensor to place the end-effector at the optimal position for manipulation by searching for salient features of the object. 

Tasks typically start with a Cartesian motion. First, the arm must approach the appropriate object specified by the task, so once the perception node gives the pose estimate of an object, the Cartesian control makes use of inverse kinematics to plan a trajectory and then the plan executes to appropriately position the arm. Note that the position is specified as offsets and rotations from an object centroid based-off manual experimentation. Once executed, the task goes into search mode to get in position to grasp the object properly and then closes the hand. If the task requires further large scale movements, i.e., move a spoon to a bowl, then the Cartesian control mode will be activated again. 
% * <brro5352@colorado.edu> 2017-01-19T21:49:25.509Z:
% 
% > which we will discuss in more detail later, 
% Not sure if I have enough to talk about search. It moves back and forth tell it finds something
% 
% ^ <brro5352@colorado.edu> 2017-01-19T21:55:59.094Z.

Limitations in the perception system, due to noise from the RGB-D sensor and miscalibration, lead to uncertainty in the object's pose. Because of this uncertainty, exclusively relying on open-loop position control may lead to collisions or failed execution of the task, for example failing to grasp a spoon because it is not within reach. So to deal with this uncertainty in perception the Cartesian control positions the arm at a safe offset from the feature of interest, and then use velocity control to search for a task-relevant feature, for example the handle of a spoon. Once the feature is detected, which we will discuss in more detail below, the system will proceed with the appropriate action such as grasping the object or pushing the object. If the object is not found during the search, the sub task is restarted.

\subsubsection{Sensor Feedback}

\begin{figure}
\centering
\includegraphics[width=12cm, height = 8cm, keepaspectratio]{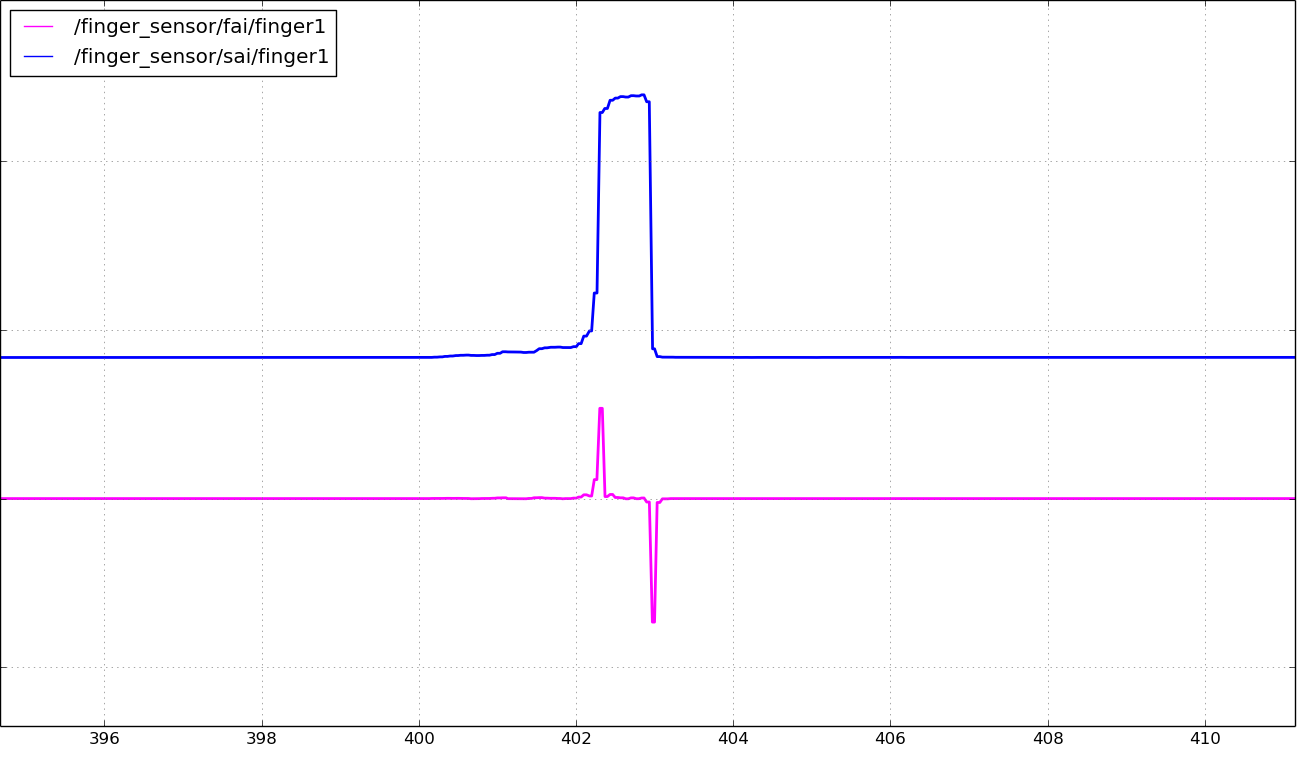}
\caption{Sensor values (analog reading) versus time for the SA-I (blue) and FA-I (pink) channel equivalents from the 1st finger on the Jaco arm. The gradual increase in the SA-I channel refers to an object detection event. The first peak in the FA-I channel refers to the contact event. A drop in the SA-I channel refers to the object separation event. The second down peak in the FA-I channel is the release event. \label{fig:snrOutput}}
\end{figure}

We use two distinct channels of information from the finger sensors (proximity and contact) within our feedback controller. Passing the non-linear sensor input through a high-pass filter with 20 Hz cut-off frequency \cite{patel2016b} allows us to detect contact, which appears as extrema in the high-pass signal. The resulting signals are roughly equivalent to the SA-I and FA-I signals in the human hand, that is constant pressure and dynamic tactile events, respectively  \cite{patel2016b}. After calibrating the sensors by fixing the base value of non-linear and surface dependent sensory input moments before executing the grasp, values ranging above and below specific thresholds are considered object and contact detection events respectively (Figure \ref{fig:snrOutput}). The pseudo code for both these event detection is provided in Algorithms \ref{touchdetection} and \ref{objectdetection}.   
% * <brro5352@colorado.edu> 2017-01-19T20:57:51.542Z:
% 
% > assing the non-linear sensor input through a high-pass filter with 20 Hz cut-off frequency allows us to detect contact, which appears as extrema in the high-pass signal. 
% Too much detail?
% 
% ^ <brro5352@colorado.edu> 2017-01-19T21:43:07.240Z.
% * <brro5352@colorado.edu> 2017-01-19T20:57:16.950Z:
% 
% > feedback controller.
% Should we just say finger sensor
% 
% ^.

\begin{algorithm}
\caption{Touch detection}\label{touchdetection}
\begin{algorithmic}[1]
% \State $\textit{current\_fingers\_touch} \gets \text{[False]*3 }$ \Comment{$\oplus$: bitwise exclusive-or}
\Function{detect\_touch}{$current\_FAI\_finger, $y}
  \State $\textit{touch} \gets \textit{current\_fingers\_touch}$
  \State $\textit{FAI} \gets \text{[current\_FAI\_finger1, current\_FAI\_finger2, current\_FAI\_finger3]}$
  \For{$fingers \gets 1 \textrm{ to } 3$}
    \If{$FAI[fingers] < -threshold$ \& $current\_finger\_touch == False$}
      \Let{$touch[fingers]$}{$True$}
        \EndIf
    \If{$FAI[fingers] > threshold$ \& $current\_finger\_touch == True$}
      \Let{$touch[fingers]$}{$False$}
        \EndIf
      \EndFor
    \EndFunction

\end{algorithmic}
\end{algorithm}

\begin{algorithm}
\caption{Object detection}\label{objectdetection}
\begin{algorithmic}[1]
% \State $\textit{current\_object\_detect} \gets \text{[False]*3 }$ \Comment{$\oplus$: bitwise exclusive-or}
\Function{detect\_object}{$current\_SAI\_finger, $y}
  \State $\textit{detected} \gets \textit{current\_object\_detect}$
  \State $\textit{SAI} \gets \text{[current\_SAI\_finger1, current\_SAI\_finger2, current\_SAI\_finger3]}$
  \For{$fingers \gets 1 \textrm{ to } 3$}
    \If{$SAI[fingers] < -thershold$ \& $current\_object\_detect == False$}
      \Let{$detected[fingers]$}{$True$}
        \EndIf
    \If{$FAI[fingers] > threshold$ \& $current\_object\_detect == True$}
      \Let{$detected[fingers]$}{$False$}
        \EndIf
      \EndFor
    \EndFunction
\end{algorithmic}
\end{algorithm}

\section{Results}

In this section we describe how the finger sensors and perception pipeline facilitated grasping and manipulation using object recognition, contact point detection, and pose estimation for the ten competition tasks. Combining 3D perception with proximity information greatly increased the robustness of our manipulation approach by mitigating calibration error and sensor noise. 

In \emph{task i and ii} both required perceiving the thin and narrow spoon handle. Without using the finger sensors, failure modes include positioning the end-effector to far away from the spoon or running into the spoon and thereby changing its position. The proximity information from the sensors enabled us to position the end-effector correctly in a position to properly grasp the spoon using hard-coded search routines around the estimated position. Once the hand was in a position to grasp the spoon, it was moved to make contact with the spoon. If the robot continued to move after initial contact with the spoon, the spoon could get displaced leading to an empty grasp. The contact/release information from the sensors indicated when the fingers made contact with the spoon, and terminated the motion of the hand in a timely manner. The spoon was then securely grasped by closing the fingers in a controlled manner.  The following tasks were then straightforward to execute via prerecorded motions; \emph{task i} required simple motions to scoop peas and deposit them and \emph{task ii} required stirring of the contents in a cup. With a proper orientation of the spoon after grasping, both tasks were easily completed. 

Grasping a straw out of a cup (\emph{task vi}) was similar to grasping the spoon. Using the proximity information from the sensors we could correctly locate the straw in space. The sensor's high sensitivity allowed us to identify the touch event before the grasp started to displace the straw and successfully pick it up. It was difficult, however, to insert the straw into the plastic cup through the small opening in the lid due to the comparably large error in perception (3-5cm), and we did not use an additional step to use the sensors to properly locate \cite{cox2017} the cup.

Unlike the aforementioned tasks, \emph{task iii}, grasping and shaking a salt dispenser, was trivial in terms of perception and grasping. Dynamic manipulation, on the other hand, proved difficult for the Kinova robot. Sufficient jerk to release salt from the shaker could not be achieved within the limits of the arm. Here, using wrist rotation instead of moving the entire arm led to best results, but still dispensed the salt at a very slow rate that made the task take a long time to complete.  

For \emph{task iv}, the Kinova hand was able to create sufficient force closure with the USB light to pull it out of a USB connector in the socket. Plugging the connector back in was difficult due to lack of stiffness in the hand and the light itself, which was made from a flexible material. Solving this task successfully requires grasping the light as close as possible to its stiffest part and then use repeated trial and error or additional optical sensing.

Picking up a hammer and punching nails in a foam block (\emph{task v}), was challenging due to the weight of the hammer and lack of stiffness in the Kinova hand. We note that since using in-hand sensors to make up for uncertainty, none of the tasks took advantage of the built-in compliance of the Kinova hand. 
% * <brro5352@colorado.edu> 2017-01-20T20:19:41.944Z:
% 
% > none of the tasks took advantage of the built-in compliance of the Kinova hand. 
% Can you clarify this a bit.
% 
% ^.

% * <brro5352@colorado.edu> 2017-01-20T20:20:38.305Z:
% 
% > Grasping a straw out of a cup (\emph{task vi}) was similar to grasping the spoon. 
% We say this more elegantly, like we used an approach that is identical to the one used for the spoon task.
% 
% ^.

Inserting a screwdriver into a nut (\emph{task vii}) again emphasized precision. Picking up the screwdriver was relatively simple, however correctly inserting the driver into the nut was not possible with our setup. Although trial and error based on an initial estimate on the nut's pose is a viable strategy, the nut does not have a large enough area that is suitable for self-alignment. In addition, the rotation of the screwdriver is crucial to catching the nut in order to apply a rotational force. The limited resolution in our perception pipeline does not provide us with enough information to align these items properly for manipulation.  

Similar to removing the USB night light, charging and emptying a syringe with air (\emph{task ix}) was a test to the robot hand's ability to apply a pinch grasp strongly. Since a task like this requires two arms to perform, participants with a single robot arm were allowed to have a teammate hold the syringe with their hand while the robot pulled the syringe handle.

Picking up a towel and hanging it onto a hanger (\emph{task iv}) was straightforward as it was supposed to simply picked and placed. Here, the challenge was picking up the towel very close to the table surface. Proximity information in the fingers allowed us to stop the arm above the table at a distance which was safe enough for the fingers not to brush against the table and reliable enough to grab the towel.

The final and most difficult task was taking a pair of scissors and cutting a paper along predefined lines (\emph{task viii}). One had to first identify the lines on the paper which we did using a standard line-detection algorithm. Picking up the scissors was facilitated with the handle hanging over the table. The challenging part was orienting the scissors correctly to cut along the lines. The fingers of the Jaco arm did not have the ability to comply with the shape of the scissors when opening it (i.e., bending the fingers such that the hand does not lose grip of the scissors while repeatedly opening and closing the scissors). The scissors hence loose contact with the fingers when either opening or closing, making this task impractical with the gripper configuration used. 

We have focused exclusively on the manipulation aspect of the competition as the bin-picking task would require a different perception strategy, focusing on object identification. 

\section{Discussion}
A key insight in addressing a wide variety of tasks in a competitive environment was that 3D perception, mechanical compliance, and tactile sensing complement each other and deficiencies in one can be made up by the other to some extent. Indeed, many teams were able to solve a majority of the tasks without using any perception, but relied exclusively on mechanical compliance and hard coded positions of objects. Analogously, humans might be able to perform tasks without tactile sensing or being blind-folded, but it is the combination of the two that makes them most efficient. 

Indeed, better 3D perception and calibration might have allowed us to forgo tactile sensing altogether. Likewise, some of the tasks could have been accomplished using exclusively in-hand proximity and contact sensing. It might be this redundancy, which lets the community mostly focus on thoroughly exploring single sensing modalities rather that exploring comprehensive solutions that combine 3D perception, tactile sensing and mechanical compliance. 

We also learned valuable lessons in how to specify competition rules in order to push the community toward generalizable outcomes. Bin picking and tabletop manipulation are indeed sufficiently different problems that the system presented here was not able to solve tasks in bot categories, albeit mainly due to different requirements in perception. A loop-hole in this year's competition rules was that augmenting the objects was not explicitly forbidden, allowing one team to mount foam cubes onto individual objects that could be grasped by the Baxter robot's standard gripper with a large margin of error. This is an interesting solution, which uses compliance in a smart way and would lead to acceptable outcomes in some constraint scenarios, but only poorly generalizes to household manipulation tasks. 

%Some teams approached the Pick-and-Place stage of this track with a random search motion of the arm in the basket. Objects were searched in random fashion in the basket and a successful grasp of an object was finished by placing it in the desired location on the table. Team Dorabot\&Cobot finished this task quite smoothly. Considering the limited aperture of their Baxter robot the team from Tsinghua university glued small foams cubes to the objects and grasped them instead of the objects itself. Through kinesthetic teaching they recorded the entire motion of the arm, from grasping the objects(foam cubes) to putting them in their correct orientation and location on the table and then just replayed the motion. Modifications to objects and considering kinesthetic teaching as full autonomy they received a full score in this stage. 

As in the Amazon Picking Challenge \cite{correll2016analysis}, proximity and tactile sensing were underrepresented in the competition. Albeit we greatly benefited from the availability of contact and touch information, all of the tasks could be solved relying on accurate pose estimation and compliance. The limitations of this approach are best illustrated in the towel manipulation tasks. Here, most teams let their robot's hands run into the table in order to make sure they are close enough to the towel. While this worked for this task, the force exerted by compliant robots might lead to undesired outcomes in some environments, and excessive use of such strategies is unlikely in future real world applications. 

%For task-6 none of the participants had an advance perception and control techniques that could located the hole and manage to insert the straw straight into the cup. The organizers thus had slightly changed the rules for this task and had allowed for a manual control of the arm at the final stage for inserting the straw.

%In another task, participants whose robot gripper's pinch force was strong enough could unplug the USB lights from the socket easily, however inserting it back was difficult for all teams. Albeit compliance usually helps in peg-in-hole like tasks, a successful insertion requires more precise sensing than is possible with a RGB-D camera.

Some of the tasks demonstrated the need for dynamic control strategies. Specifically, position and velocity-based controllers are not sufficient for tasks like emptying the salt shaker, which require accurate control of jerk. Similarly, undoing a plug leads to significant jerk, which leads to disturbance of the environment. The requirements on dynamic control are therefore two-fold: first, the ability to specify not only position and velocity, but also acceleration profiles. Second, high-bandwidth impedance control, usually available only in expensive industrial robot arms, is not a luxury, but safety critical for operations with quickly changing loading conditions. 

%Here, it is not sufficient to design the mechanism as stiff as possible. For example, almost all the robot hands, including ours, could not hold the scissors properly due to lack of appropriate compliance and stiffness component. Except the UNIPI-IIT-QB team from Italy who showed some promising scissor manipulation skills with their PISA/IIT Softhand in track1 of the competition but again the question is concerning the intergrability and control simplicity with a robotic arm. No team could finish the task-8 in its entirety.

%The wining team from Tsinghua University scored a total of 165/300 in the autonomous track of the competition with team Dorabot\&Cobot right behind them with 155/300.

%As with the Amazon Picking Challenge \cite{correll2016analysis}, it turns out that there exist a wide variety of solutions that are capable of solving the proposed tasks within their specified constraints by trading perception and planning with mechanism design. Assuming the proposed tasks to be of commercial relevance and to completely represent the problem, such an approach is viable to advance our knowledge and find the ``best'' solution. In practice, however, winning solutions often only poorly generalize. For example, the dominant solution in the picking challenge, suction, would require to modify the packaging of items that cannot be grasped that way, which might be feasible in a warehousing environment, yet cannot be expected to transfer to a household environment. Similar reasoning applies to assumption on localization priors or specific object shapes. 

The largest source of error resulted from errors in calibration. These include the intrinsic camera parameters, but also finding the transformation and rotation from the ASUS Xtion to the base of the arm. While there exist more powerful calibration strategies than chosen here and we could also permanently mount the camera to the robot's arm frame, we note that different tasks require different camera perspectives. We therefore consider calibration an open problem, and are interested in exploring solutions that augment object localization and pose calibration using tactile sensing \cite{cox2017}, as well as using the 3D model of the robot itself to add data points to the calibration process. 

%To remove inconsistencies in calibration for successive trials we plan to build a camera jig attached to the arm base. With the known location of the camera in the world frame and the available transform between the camera and the arm base frame we will be able to apply this transformation in each new trial to the camera frame for calibration and align it nearly perfectly with the world frame.

All of the tasks in this competition could be solved without using any motion planning. That is, all motions were executed by simply commanding the robot to a Cartesian pose, assuming that there exist a collision-free trajectory. As this cannot be assumed in a real-world application, we plan to integrate the solution presented here with the motion planning framework MoveIt! \cite{coleman14b}. This task is less straightforward than it sounds as the discrete planning approach that is customary in motion planning does not smoothly integrate with continuous feedback control, and how to do this properly is subject to further research.

%To improve the control of the arm we plan to integrate Moveit! in our grapsing pipeline. Apart from planned motion trajectories the library also provides various other facilities like collision checking. We thus anticipate the arm to no longer collide with the table as a result of an off-calibration by a few centimeters. Object attachment and detachment facilities will help us have a better control over the objects and avoid collisions of object with the environment. A knowledge of the object's orientation with respect to arm will help with better manipulation of the object. For instance in tasks like unplugging and plugging a USB light. 
%A more distant future plan is to integrate OpenRAVE in our grasping pipeline to design and test robotic hands in simulation and also take prior knowledge of grasps into account by ranking the grasps. 

\section{Conclusion}
We have presented a comprehensive perception and manipulation pipeline that combines 3D perception with proximity and tactile sensing using exclusively commercially available hardware. All software developed for this project is available open-source\footnote{\url{https://github.com/correlllab/cu-perception-manipulation-stack}} and continues to be expanded on.

We have shown that in-hand proximity and tactile sensing can dramatically improve the robustness of a large variety of grasping and manipulation tasks in face of uncertainty in sensing and actuation, and we argue that those sensing modalities are critical for performing robust manipulation in the real world. 

Challenges that remain towards this end are: (1) increasing the accuracy of orientation estimation of objects and the efficiency of 3D perception for larger data sets of objects, (2) better integration of deliberative and reactive control strategies, and (3) improved mechanism design allowing for controlling compliance and stiffness to be able to manipulate heavy objects as well as those that require deformation of the hand.

\bibliographystyle{splncs03}
% \nopagebreak
\bibliography{imprvGrasp}
\end{document}